\documentclass[sigconf,screen,nonacm]{acmart}

\AtBeginDocument{%
  \providecommand\BibTeX{{%
    \normalfont B\kern-0.5em{\scshape i\kern-0.25em b}\kern-0.8em\TeX}}}

\usepackage{subfigure}
\usepackage{multirow}

\begin{document}

\title{Assessing unconstrained surgical cuttings in VR using CNNs}

\author{Ilias Chrysovergis}
\orcid{0000-0002-5434-2175}
\email{ilias.chrysovergis@oramavr.com}
\affiliation{%
  \institution{ORamaVR}
  \country{Greece}
}

\author{Manos Kamarianakis}
\email{kamarianakis@uoc.gr}
\orcid{0000-0001-6577-0354}
\affiliation{%
  \institution{FORTH - ICS, University of Crete, ORamaVR}
  \country{Greece}
}

\author{Mike Kentros}
\orcid{0000-0002-3461-1657}
\email{mike.kentros@oramavr.com}
\affiliation{%
  \institution{FORTH - ICS, University of Crete, ORamaVR}
  \country{Greece}
}

\author{Dimitris Angelis}
\orcid{0000-0003-2751-7790}
\email{dimitris.aggelis@oramavr.com}
\affiliation{%
  \institution{University of Crete, ORamaVR}
  \country{Greece}
}

\author{Antonis Protopsaltis}
\orcid{0000-0002-5670-1151}
\email{aprotopsaltis@uowm.gr}
\affiliation{%
  \institution{University of Western Macedonia, ORamaVR}
  \country{Greece}
}

\author{George Papagiannakis}
\orcid{0000-0002-2977-9850}
\email{papagian@ics.forth.gr}
\affiliation{%
  \institution{FORTH - ICS, University of Crete, ORamaVR}
  \country{Greece}
}

\renewcommand{\shortauthors}{Chrysovergis, Kamarianakis, Kentros et al.}


\maketitle

\section{Introduction}
In the recent years, industry and academia \cite{review_vr_training_applications} have massively adopted 
Virtual Reality (VR) applications to train students and personnel. 
Despite the effort, only limited systems involve procedures for 
assessing user progress inside the immersive environment, that 
either evaluate only trivial tasks or require a huge amount of 
time by the reviewers \cite{Southgate2020}.
On the other hand, the need for real-time automated evaluation 
of user's actions is constantly increasing.
State-of-the-art methods for similar tasks either require the 
development of complicated task specific computer vision algorithms
\cite{Zia2016} or support very simple tasks \cite{Lahanas2015}.

This work proposes a deep learning based system, that is able 
to assess, in real-time, user actions within a VR training scenario. 
The method enables the rapid development of trained assessment 
functions, since it utilizes data augmentation to minimize the 
amount of labelled data that need to be collected. 
Furthermore, by using transfer learning, these assessment functions 
can be reconfigured to support  similar tasks, thus reducing even 
more the amount of training data.
In this paper, we present the results of our method for the task of 
tearing a deformable 3D model \cite{Kamarianakis_Papagiannakis_2021}.

Different machine and deep learning algorithms 
\cite{wangTimeSeriesClassification2016, dempsterROCKETExceptionallyFast2020} 
were considered and compared (see Table~\ref{tbl:comparison}). 
Ultimately, our proposed model is a Convolutional Neural Network 
(CNN), trained on a dataset created with a data augmentation technique
\cite{iwanaTimeSeriesData2020}.

\begin{figure}[b]
    \centering
    \includegraphics[height=60mm, width=0.47\textwidth]{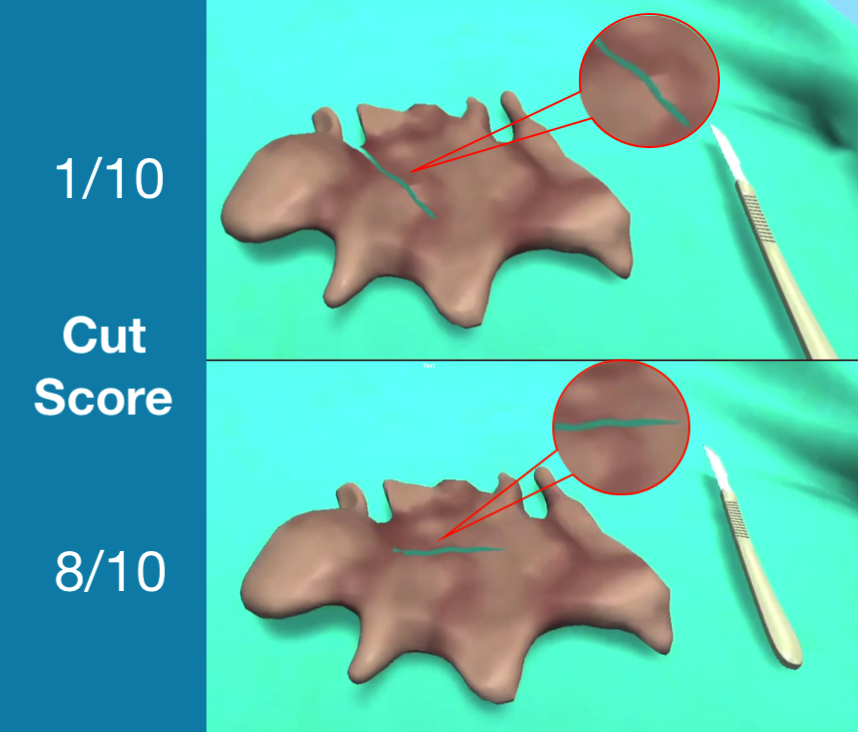}
    \caption{The trained Convolutional Neural Network assesses 
    the cutting action performed on a deformable 3D model.}
    \Description{The trained Convolutional Neural Network assesses 
    the cutting action performed on a deformable 3D model.}
    \label{fig:scores}
\end{figure}

\section{Our Approach}

\textbf{\textit{Data Collection:}} 
The input datasets, generated by multiple execution of the 
training tasks, are labelled by a score, specified by the task 
designer, in our case a surgeon.
These datasets contain per-frame captured transformation data 
(translation \& rotation) of the active virtual tool  (e.g. a scalpel), 
forming its trajectory and representing the user gesture on each action.

\textbf{\textit{Data Sampling and Augmentation:}}
Since the execution of an action is user dependent, the required 
task completion times vary and the generated trajectory lengths $M$ 
are non-uniform.
To amend this, random samples are taken, creating $N$-length 
trajectories that can be fed to the CNN. The remaining data of 
the long trajectory are sampled to create more $N$-length 
trajectories, augmenting the dataset. This technique significantly 
increases the number of the training data and therefore enhances 
the training process.

\textbf{\textit{Neural Network (NN) Architecture:}}
Since low training and inference times are preferred, a lightweight 
model, able to provide high-accuracy results, was created. 
This was achieved by using a $15$-layer CNN, which consisted of $4$ 
sets of two Convolutional, two ReLU, and a Batch Normalization 
layer, along with $3$ final layers, a Global Average Pooling, 
a Flatten and a Dense layer, which formed the classifier.
Each of the $4$ sets produces an output of higher dimensionality.
Thus, as the input is passed down deeper in the network, more 
complex patterns in the trajectory are found.

\textbf{\textit{Training:}}
After being collected and sampled, the data are split randomly 
in training, validation and testing sets.
The training set is used to train the NN, while the validation 
set is used to find the best hyper-parameters and prevent over-fitting. 
Lastly, the testing set is used to evaluate the model's accuracy.
The NN was trained using the Adam optimizer with a learning rate 
of $10^{-4}$,  the sparse categorical cross entropy loss function, 
with
100 epochs and a batch size of 16.

\textbf{\textit{Inference:}}
The trained NN model is then imported in the user's actions 
assessment module of the VR application.
As before, these actions are captured and sampled into smaller 
length trajectories, which are fed into the model.
The user's score is produced instantly with no overhead on the 
application's performance.

\textbf{\textit{Retraining:}}
The proposed NN can be easily adapted for similar tasks, utilizing 
transfer learning.
In this respect, the first layers of the NN are frozen while the 
deepest ones are retrained using a small amount of new training data.
This procedure reduces training time and the required amount of 
the collected data, and provides high accurate results, since the 
model's first layers represent more generic patterns, that are 
common to similar actions. 

\balance
\textbf{\textit{Scoring System:}}
The user's actions are evaluated with a score in the range of $0-10$.
However, the NN regards scores $0-5$ ($6$ classes), which requires 
less training data.
The model outputs the probabilities $p(i)$ of a trajectory belonging 
to each of the 6 classes. Ultimately,  the score $S_{tr}(k)$ for a 
single trajectory $k$ (in range $0-10$) and the final score $S_{tot}$ 
for the entire action are obtained by evaluating
\begin{equation*}
    S_{tr}(k) = 2\sum_{i=0}^{5} i\cdot p(i) \ \  \text{ and }\ \ \  S_{tot} = \frac{1}{K}\sum_{k=0}^{K}S_{tr}(k).
\end{equation*}

\begin{table}[tb]
    \centering
    \begin{tabular}{|c||c|c|c|c|}
        \hline
        No. of & Logistic & \multirow{2}{*}{KNN} & \multirow{2}{*}{SVM}  & \multirow{2}{*}{CNN}   \\
        Classes & Regression & & & \\
        \hline\hline 
        2 & $100\%$ & $100\%$ & $100\%$ & $100\%$\\ \hline
        3 & $100\%$ & $100\%$ & $100\%$ & $100\%$\\ \hline
        6 & $39\%$ & $66\%$ & $83\%$ & $95\%$\\ \hline
    \end{tabular}
    \caption{Table comparing the accuracy 
    of traditional ML techniques and the proposed CNN (last column) for different number of classes.}
    \Description{Table comparing the accuracy 
    of traditional ML techniques and the proposed CNN (last column) for different number of classes.}
    \label{tbl:comparison}
\end{table}

\begin{figure}[ht]
    \includegraphics[width=0.285\textwidth]{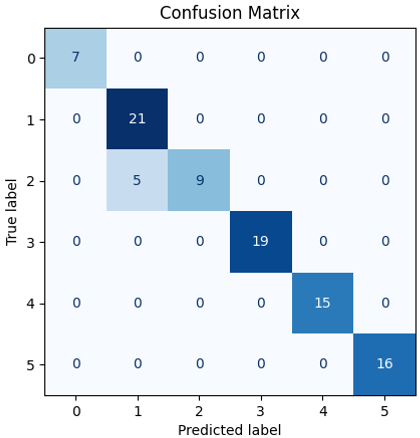}
    \Description{The confusion matrix of the convolutional neural network.}
    \caption{The confusion matrix of the convolutional neural network for the $6$ classes.}
    \label{fig:conf_matrix}
\end{figure}

\section{Results \& Discussion}
Figure \ref{fig:scores} shows the  scores calculated by the DL model 
for two different cuts performed by a user on a deformable 3D model. 
The action graded with 10 during training was collinear and of greater 
length to the one that scored 8. 
In Table~\ref{tbl:comparison}, a comparison of the accuracy of different 
models is presented for various number of classes. 
In the case of $2$ or $3$ classes, all models perform correctly, 
however, when $6$ classes are used, only our proposed CNN method 
achieves an acceptable performance. 
In Figure~\ref{fig:conf_matrix}, the confusion matrix of the trained 
model depicts that, for most classes, the true labels coincide with 
the predicted ones. The only exception is the third label which 
sometimes is predicted falsely as the second one, due to the fact 
that the specific classes have very similar trajectories, making it 
difficult for the model to distinguish among them. 
As this error could also be made by a human, the results are 
considered satisfying.
However, we plan to mitigate that issue by  performing semi-supervised 
learning \cite{semi_supervised_learning} using unlabelled data 
collected by users executing cutting actions using the MAGES 
SDK \cite{papagiannakis2020mages}.
With such training data, the model will learn to better distinguish 
among different trajectories, allowing the classifier to achieve 
better performance for all classes.

\begin{acks}
The project was partially funded by the European Union’s 
Horizon 2020 research and innovation programme under grant agreements 
No 871793 (ACCORDION) and No 101016509 (CHARITY).
\end{acks}

\bibliographystyle{ACM-Reference-Format}
\bibliography{references}

\end{document}